\documentclass[11pt]{article}    
\usepackage{hyperref}
\usepackage{url}

\usepackage{xcolor}
\usepackage{graphicx}
\usepackage{fancyvrb}
\usepackage{url}
\usepackage{paralist}
\usepackage{amsmath,amsfonts,amssymb}
\usepackage{booktabs}
\usepackage{multirow}

\usepackage{geometry}
\usepackage{graphicx}   
\usepackage{color}
\usepackage{xcolor}
\usepackage{amssymb}
\usepackage{hyperref}
\usepackage{authblk}

\title{The Second Conversational Intelligence Challenge (ConvAI2)}

\author[1]{Emily Dinan}
\author[2]{Varvara Logacheva}
\author[2]{Valentin Malykh}
\author[1]{Alexander Miller}
\author[1]{Kurt Shuster}
\author[1]{Jack Urbanek}
\author[1]{Douwe Kiela}
\author[1]{Arthur Szlam}
\author[3]{Iulian Serban}
\author[4,1]{Ryan Lowe}
\author[5]{Shrimai Prabhumoye}
\author[5]{Alan W Black}
\author[5]{Alexander Rudnicky}
\author[6]{Jason Williams}
\author[1,4]{Joelle Pineau}
\author[2]{Mikhail Burtsev}
\author[1]{Jason Weston}

\affil[1]{Facebook AI Research}
\affil[2]{Moscow  Institute  of  Physics  and Technology}
\affil[3]{University of Montreal}
\affil[4]{McGill University}
\affil[5]{Carnegie Mellon University}
\affil[6]{Microsoft Research}

\date{}

\begin{document}

\maketitle


\begin{abstract}
We describe the setting and results of the
ConvAI2 NeurIPS competition that aims to further 
the state-of-the-art in open-domain chatbots. 
Some key takeaways from the competition are: (i) pretrained Transformer variants are currently the best performing models on this task, (ii) 
but to improve performance on multi-turn conversations with humans,
future systems must go beyond single word metrics like perplexity to measure the performance
across sequences of utterances (conversations) – in terms of repetition, consistency and
balance of dialogue acts (e.g. how many questions asked vs. answered).

%
\end{abstract}

\section{Overview of the competition}

The Conversational Intelligence Challenge\footnote{\url{http://convai.io/}} aims at finding approaches to creating high-quality dialogue agents capable of meaningful open domain conversation. Today, the progress in the field is significantly hampered by the absence of established benchmark tasks for non-goal-oriented dialogue systems (chatbots) and solid evaluation criteria for automatic assessment of dialogue quality.
The aim of this competition was therefore to establish a concrete scenario for testing chatbots 
that aim to engage humans, and become a standard evaluation tool in order to make such systems directly comparable, including open source datasets, evaluation code (both automatic evaluations and code to run the human evaluation on Mechanical Turk), model baselines and the winning model itself.


This is the second Conversational Intelligence (ConvAI) Challenge; the previous one was conducted under the scope of NeurIPS 2017 Competitions track. 
Taking into account the results of the previous edition, this year we improved the task, the evaluation process, and the human conversationalists' experience. We did this in part by making the setup simpler for the competitors, and in part by making the conversations more engaging for humans. We provided a dataset from the beginning, {\sc Persona-Chat}, whose training set consists of conversations between crowdworkers who were randomly paired and asked to act the part of a given provided persona (randomly assigned, and created by another set of crowdworkers). The paired workers were asked to chat naturally and to get to know each other during the conversation. This produces interesting and engaging conversations that learning agents can try to mimic. 
The {\sc Persona-Chat} dataset is designed to facilitate research into alleviating some of the issues that traditional chit-chat models face, and with the aim of making such models more consistent and engaging, by endowing them with a persona \cite{zhang2018personalizing}.
Models are thus trained to both ask and answer questions about personal topics, and the resulting dialogue can be used to build a model of the persona of the speaking partner.

Competitors' models were compared in three ways: (i) automatic evaluation metrics on a new test set hidden from the competitors; (ii) evaluation on Amazon Mechanical Turk; and (iii) `wild' live evaluation by volunteers having conversations with the bots. 
We declared winners in the automatic evaluation tracks,  but the grand prize was awarded to the best performing system in human evaluations.

The winner in the automatic evaluation tracks by a significant margin was the team Hugging Face, 
however the grand prize winner from human evaluations was Lost in Conversation (Hugging Face coming in second place, with 23 entrants in total)\footnote{The Lost in Conversation entry will be described in detail in separate publication by their team.}. 
There are a number of key takeaways from our analysis of the results, indicating that the automatic evaluations show some correlation to human evaluations, but fail to take into account important aspects of multi-turn conversation that humans consider important, in particular the balance of dialogue acts throughout the conversation (e.g. the amount of questions asked versus answered).


\subsection{Previous competitions and task formulation}

There have been a number of competitions on question answering (e.g. quiz bowl) which can be seen as single-turn goal-directed dialogue, as well as competitions on goal-directed dialogue involving dialogue state tracking (including 5 iterations of the DSTC challenge), e.g. for booking restaurants or tourist information.
Those do not explicitly address the ``chit-chat'' setting of dialogue about general topics which is not goal-directed, although later DSTC challenges do address chit-chat.

The first edition of the Conversational Intelligence Challenge took place at the NeurIPS 2017 Competition track in the form of a live competition. The task was for an agent to carry out intelligent and natural conversations about specific snippets from Wikipedia articles with humans, which was not engaging to all human participants.

Ten dialogue systems participated in the 2017 competition. The majority of them combined multiple conversational models such as question answering and chit-chat systems to make conversations more natural. The evaluation of chatbots was performed by human assessors. More than 1,500 volunteers were attracted and over 4,000 dialogues were collected during the competition. All the data and the solutions of the winners are available via the competition repo.\footnote{\url{http://convai.io/2017/data/}}\textsuperscript{,}\footnote{\url{https://github.com/DeepPavlov/convai/tree/master/2017/solutions}} The final score of the dialogue quality for the best bot was 2.746 compared to 3.8 for human. This demonstrates that current technology allows supporting dialogue on a given topic but with quality significantly lower than that of humans.

In contrast to the first edition, the 2018 competition focused on general chit-chat about people's interests, rather than on encyclopedic facts. To our knowledge, no other competition has focused on a dataset like this. Importantly, we provided a large training set and validation set in a standard setup, complete with code for baseline systems for entrants to obtain clear automatic evaluation metrics to improve upon. In the 2017 ConvAI competition, no data was initially provided but was instead collected by volunteers as the competition progressed, which may have led to fewer participants.

Outside of NeurIPS, the most similar competition is probably the Alexa Prize\footnote{\url{https://developer.amazon.com/alexaprize}}. This is a competition to build a socialbot that can converse coherently and engagingly with humans on popular topics for 20 minutes.  The top bots were selected by Amazon Alexa customers and the Amazon panel and competed head-to-head in front of three judges in November 2017. 
Another small scale analogue is  the Loebner Prize.\footnote{\url{https://en.wikipedia.org/wiki/Loebner_Prize}}
Alexa Prize data and models are not in the open domain, whereas our competition aims to have as deliverables both data and winning models and training code. Further, unfortunately, the outcome mostly confirmed that ensembles are useful in such tasks and did little to drive fundamental algorithm research.

The key differences from the the first (2017) ConvAI competition are the following:
\begin{itemize}
	\item The conversations focused on engaging the interlocutors by discussing personal interests (instead of encyclopedia articles they may not be interested in). 
	\item A training set was provided at the start of the competition, making the competition much more straightforward for participants.
   \item Evaluation included both automatic metrics, Amazon Mechanical Turk and `wild'  live volunteer conversations, making the evaluation much more complete. 
\end{itemize}

\begin{table}[t]
  \begin{center}
    \begin{small}
      \begin{tabular}{l|l}
        \toprule
        \textbf{Persona 1} & \textbf{Persona 2}\\
        \midrule
I like to ski & I am an artist\\
My wife does not like me anymore & I have four children\\
I have went to Mexico 4 times this year & I recently got a cat \\
I hate Mexican food &  I enjoy walking for exercise \\
I like to eat cheetos &  I love watching Game of Thrones\\
\bottomrule
\multicolumn{2}{l}{ }\\
\multicolumn{2}{l}{[PERSON 1:] Hi}\\
\multicolumn{2}{l}{[PERSON 2:] Hello ! How are you today ?}\\
\multicolumn{2}{l}{[PERSON 1:] I am good thank you , how are you.}\\
\multicolumn{2}{l}{[PERSON 2:] Great, thanks ! My children and I were just about to watch Game of Thrones. }\\
\multicolumn{2}{l}{[PERSON 1:] Nice ! How old are your children?}\\
\multicolumn{2}{l}{[PERSON 2:] I have four that range in age from 10 to 21. You?}\\
\multicolumn{2}{l}{[PERSON 1:] I do not have children at the moment.}\\ 
\multicolumn{2}{l}{[PERSON 2:] That just means you get to keep all the popcorn for yourself.}\\
\multicolumn{2}{l}{[PERSON 1:] And Cheetos at the moment!}\\
\multicolumn{2}{l}{[PERSON 2:] Good choice. Do you watch Game of Thrones?}\\
\multicolumn{2}{l}{[PERSON 1:] No, I do not have much time for TV.}\\
\multicolumn{2}{l}{[PERSON 2:] I usually spend my time painting: but, I love the show.}\\
      \end{tabular}
      \caption{Example dialogue from the {\sc Persona-Chat} dataset. Person 1 is given their own persona (top left) at the beginning of the chat, but does not know the persona of Person 2, and vice-versa. They have to get to know each other during the conversation.
 \label{table:persona-chat-example}}
    \end{small}
  \end{center}
\end{table}

\section{Competition description and set-up}

\subsection{Data}

The {\sc ConvAI2} dataset for training models is publicly available in ParlAI\footnote{https://github.com/facebookresearch/ParlAI/tree/master/parlai/tasks/convai2}, and is based on the {\sc Persona-Chat} dataset\cite{zhang2018personalizing}. See Table \ref{table:persona-chat-example} for an example dialogue. 
 The speaker pairs each have assigned profiles coming from a set of 1155 possible personas (at training time), each consisting of at least 5 profile sentences, setting aside 100 never seen before personas for validation.
 The dataset statistics are given in Table \ref{tab:convai2_stat}.

As the original  {\sc Persona-Chat} test set was released, 
we crowdsourced further data for a hidden test set unseen by the competitors for automatic evaluation. The hidden test set consisted of 100 new personas and over 1,015 dialogs.

 To avoid modeling that takes advantage of trivial word overlap,  additional rewritten sets of the same train and test personas were crowdsourced, with related sentences that are rephrases, generalizations or specializations, rendering the task much more challenging. For example ``I just got my nails done'' is revised as ``I love to pamper myself on a regular basis'' and ``I am on a diet now'' is revised as ``I need to lose weight.''

\begin{table}[h]
\centering
\begin{tabular}{l|rrr}
                & \# examples & \# dialogues & \# personas \\ \hline
Training set    & 131,438      & 17,878       & 1,155       \\
Validation set  & 7,801        & 1,000        & 100         \\
Hidden test set & 6,634        & 1,015        & 100         \\ \hline
\end{tabular}
\caption{Statistics of the ConvAI2 dataset (based on {\sc Persona-Chat}).
\label{tab:convai2_stat}
}
\end{table}

The task aims to model normal conversation when two interlocutors first meet, and get to know each other. Their aim is to be engaging, to learn about the other's interests, discuss their own interests and find common ground. 
The task is technically challenging as it involves both asking and answering questions, and maintaining a consistent persona, which is provided. 
 Conversing with current chit-chat models for even a short amount of time quickly exposes their weaknesses \cite{serban2016generative,vinyals2015neural}.
Common issues with chit-chat models  include:
(i) the lack of a consistent personality \cite{li2016persona} as they are typically trained over many dialogues each with different speakers,  (ii)
the lack of an explicit long-term memory as they are typically trained to produce an utterance given only the recent dialogue history \cite{vinyals2015neural}, and  (iii)
a tendency to produce non-specific answers like ``I don't know'' \cite{li2015diversity}. 
With this task we aim to find models that address those specific issues \cite{zhang2018personalizing}.

Note that for training, competitors were allowed to use other  additional training data as long as it was made public (or was already public). 

\subsection{Metrics}\label{sec:eval}

We first evaluated all submissions on a set of automatic metrics. 
The top 7 teams from the automatic metrics were then evaluated by humans:
\begin{itemize}
\item {\bf Automatic metrics}  - Perplexity, F1 and hits@1/20. These were computed on the hidden test. 
    \begin{itemize}
       \item Perplexity --- a metric of text fluency which is computed as $\frac{1}{m} \sum_{i=1}^{m} log p(w_i) $ for sentence  $\textbf{w}=w_1, w_2, ..., w_m$. This metric is computed only for probabilistic generative models.
        \item F$_1$-score --- $2 \cdot \frac{\text{precision} \cdot \text{recall}}{\text{precision} + \text{recall}}$. In the context of dialogue, precision is the fraction of words in the predicted response that are contained in the gold response, and recall is the fraction of words in the gold response that were in the predicted response. This can be computed for any model, retrieval-based or generative.
        
        \item Hits@1/20 --- hits@1/N is the accuracy of the next dialogue utterance when choosing between the gold response and $N - 1$ distractor responses (here, we use $N=19$). Distractor responses are random responses from the dataset. 
        Any model that can assign a score to a given candidate utterance can compute this metric. Such a method could then in principle be used in a retrieval model to score retrieved candidates.
    \end{itemize}

\item {\bf Human evaluations} - 
\begin{itemize}
    \item {\bf Amazon's Mechanical Turk:} Given the entrants' model code, we ran live experiments where Turkers chatted to a given model following instructions identical to the creation of the original dataset, but with new profiles, and then scored its performance. Performance was evaluated by asking Turkers how much they enjoyed talking to the model and having them verify which persona the model was using given the choice between the correct persona and a random one. 
    \item {\bf `Wild' Live Chat with Volunteers:} We solicited volunteers to chat to the models in a similar way to the Mechanical Turk setup. This setup was hosted through the Facebook Messenger and Telegram APIs.
\end{itemize} 
\end{itemize}

\subsection{Baselines and code available}

Source code for baseline methods for the competition were provided in the open source system ParlAI \cite{miller2017parlai}\footnote{\url{https://github.com/facebookresearch/ParlAI/tree/master/projects/convai2}}, including training loop and evaluation code.
The example models are the methods developed in \cite{zhang2018personalizing},
which we consider strong baselines. They include a retrieval-based Key-Value Memory Network, and two generative models: an LSTM-based attentive Seq2Seq model and a LSTM-based language model.




\subsection{Rules}
\begin{itemize}
    \item Competitors must provide their source code so that the hidden test set evaluation and live experiments can be computed without the team's influence, and so that the competition has further impact as those models can be released for future research to build off them. Code can be in any language, but a thin python wrapper must be provided in order to work with our evaluation and live experiment code via ParlAI's interface.
    \item Each team can only submit a maximum of once per month during the automatic metrics round.
    \item We require that the winning systems also release their training code so that their work is reproducible (although we also encourage that for all systems).
     \item Competitors should indicate which training sources are used to build their models, and whether (and how) ensembling is used.
    \item Competitors are free to augment training with other datasets as long as they are publicly released (and hence, reproducible). Hence, all entrants are expected to work on publicly available data or release the data they use to train.
\end{itemize}

\subsection{Timeline}
\begin{itemize}
    \item {\bf April 21:} Competition begins: automatic metrics leaderboard, baselines, and submission instructions are posted.
    \item {\bf May 9} Hackathon: We organized a non-compulsory hackathon around the competition: DeepHack.Chat. At the hackathon teams aimed to improve their systems,  took part in live human evaluations, and listened to lectures from researchers in the field.
    \item {\bf July 10:} `Wild' evaluation is open. Participants may submit their models to be evaluated by live volunteers.
    \item {\bf September 30:} Submissions for the automatic metrics round are closed. We invite the top seven teams from this round to prepare their submissions for the Mechanical Turk evaluation portion of the competition.
    \item {\bf December 9:} Winner of the competition is announced at NeurIPS 2018.
\end{itemize}

\subsection{Prize}

The grand prize for the winner of the human evaluations was awarded \$20,000 in funding for Amazon Mechanical Turk, in order to encourage further data collection for dialogue research. The winner in the automatic metrics received \$5,000 in AWS compute.

\section{Results and Analysis}

\subsection{Automatic Metrics}
We had over 23 teams submit models to be evaluated for the automatic metrics. The rank of each team was determined by sorting by the minimum rank of the score in any of the three metrics (F1, Hits@1, and Perplexity). The Hugging Face team performed the best in every single metric and was therefore determined to be the winner of this round. All participants and their scores on the hidden test set are shown in Table \ref{automaticleaderboard}.

The top seven teams made it to the next round. Notably, each of these teams surpassed our baseline models in some metric. The High Five team chose not to participate in the human evaluation round, so ultimately six teams participated in the next round. Refer to Section \ref{participants} for a description of the models submitted from the top-performing teams.

\begin{table}[t]
\small
\begin{center}
\begin{tabular}{l|l|l|l}
\toprule
 {\bf Team Names}    & Perplexity & Hits@1 & F1 \\
\midrule
1. Hugging Face & 16.28 & 80.7 & 19.5 \\
2. ADAPT Centre & 31.4 & - & 18.39 \\
3. Happy Minions & 29.01 & - & 16.01 \\
4. High Five & - & 65.9 & - \\
5. Mohd Shadab Alam & 29.94 & 13.8 & 16.91 \\
6. Lost in Conversation & - & 17.1 & 17.77 \\
7. Little Baby & - & 64.8 & - \\
8. Sweet Fish & - & 45.7 & - \\
9. 1st-contact & 31.98 & 13.2 & 16.42 \\ 
10. NEUROBOTICS & 35.47 & - & 16.68 \\ 
11. Cats'team & - & 35.9 & - \\
12. Sonic & 33.46 & - & 16.67 \\ 
13. Pinta & 32.49 & - & 16.39 \\ 
14. Khai Mai Alt & - & 34.6 & 13.03 \\ 
15. loopAI & - & 25.6 & - \\
16. Salty Fish & 34.32 & - & - \\
17. Team Pat & - & - & 16.11 \\ 
18. Tensorborne & 38.24 & 12.0 & 15.94 \\ 
19. Team Dialog 6 & 40.35 & 10.9 & 7.27 \\ 
20. Roboy & - & - & 15.83 \\ 
21. IamNotAdele & 66.47 & - & 13.09 \\ 
22. flooders & - & - & 15.47 \\ 
23. Clova Xiaodong Gu & - & - & 14.37 \\ 
\midrule 
Seq2Seq + Attention Baseline & 29.8 & 12.6 & 16.18 \\
Language Model Baseline & 46.0 & - & 15.02 \\ 
KV Profile Memory Baseline & - & 55.2 & 11.9 \\
\bottomrule
\end{tabular}
\end{center}
\caption{Automatic Metrics Leaderboard.
}
\label{automaticleaderboard}
\end{table}

\subsubsection{Further Analysis and Additional Automatic Metrics}

\begin{figure}
    \centering
    \includegraphics[width=0.6\textwidth]{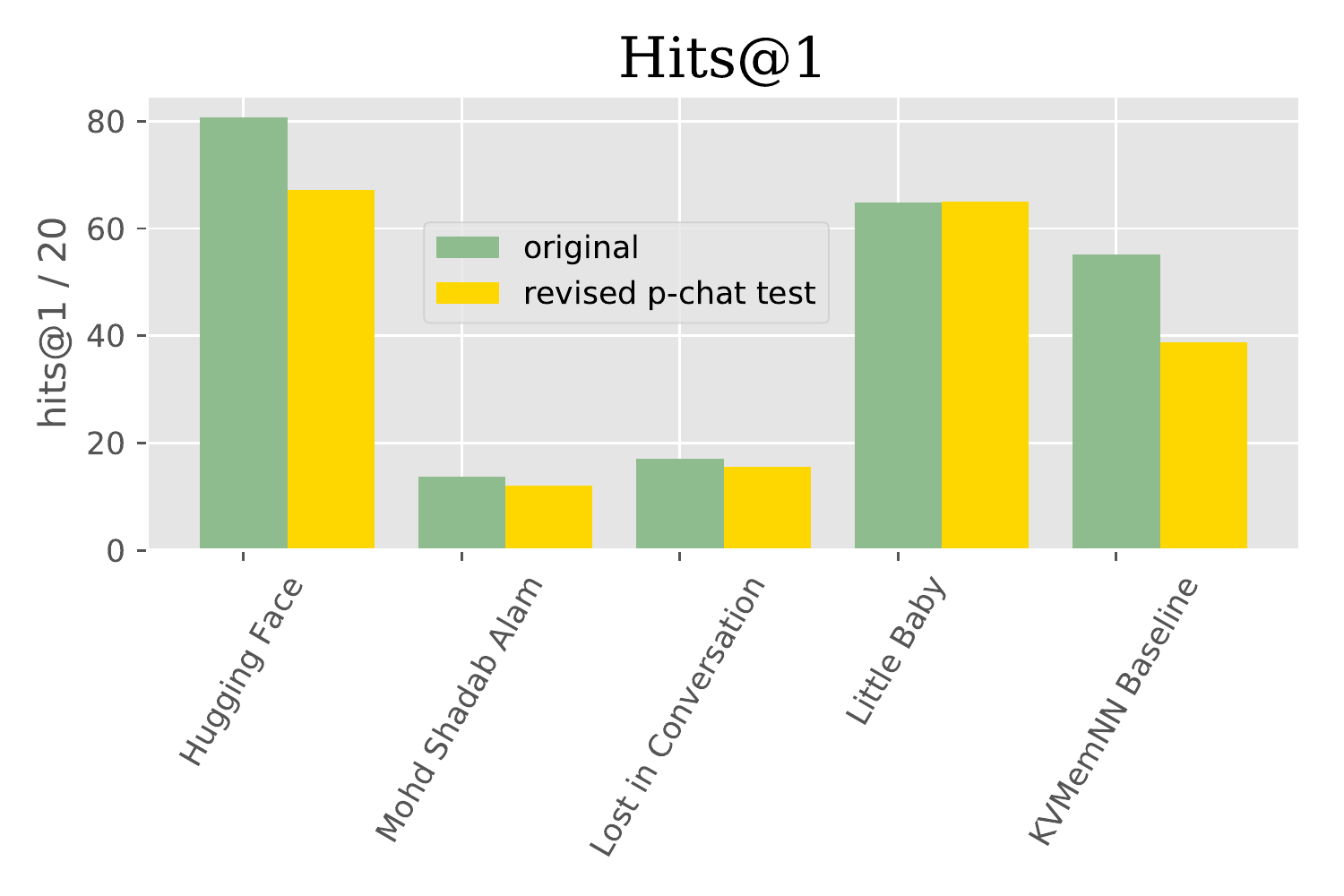}
    \caption{{\bf Revised Test Set.} Hits@1 on the revised test set vs. on the regular test set.}
    \label{fig:revised}
\end{figure}

\paragraph{Revised Personas}
We also evaluated models (from the teams in the top 7) that were capable of ranking -- i.e. models that were evaluated on the Hits@1 metric -- on the ``revised" test set.
Recall that we crowdsourced additional rewritten sets personas
as a way of measuring how much models rely on word overlap between utterances and personas for their performance, as the revised ones have little or no overlap with the original personas. 
The results are shown in Figure \ref{fig:revised}. The Hugging Face team performed the best on the revised task, with Little Baby close behind. The performance of the baseline Key-Value Memory Network baseline greatly deteriorated given the revised personas. Hence, we found the success of the best competitor's models as a good result, which we believe is due to their use of sufficient pretraining and regularization, among other factors.

\begin{figure}
    \centering
    \includegraphics[width=0.6\textwidth]{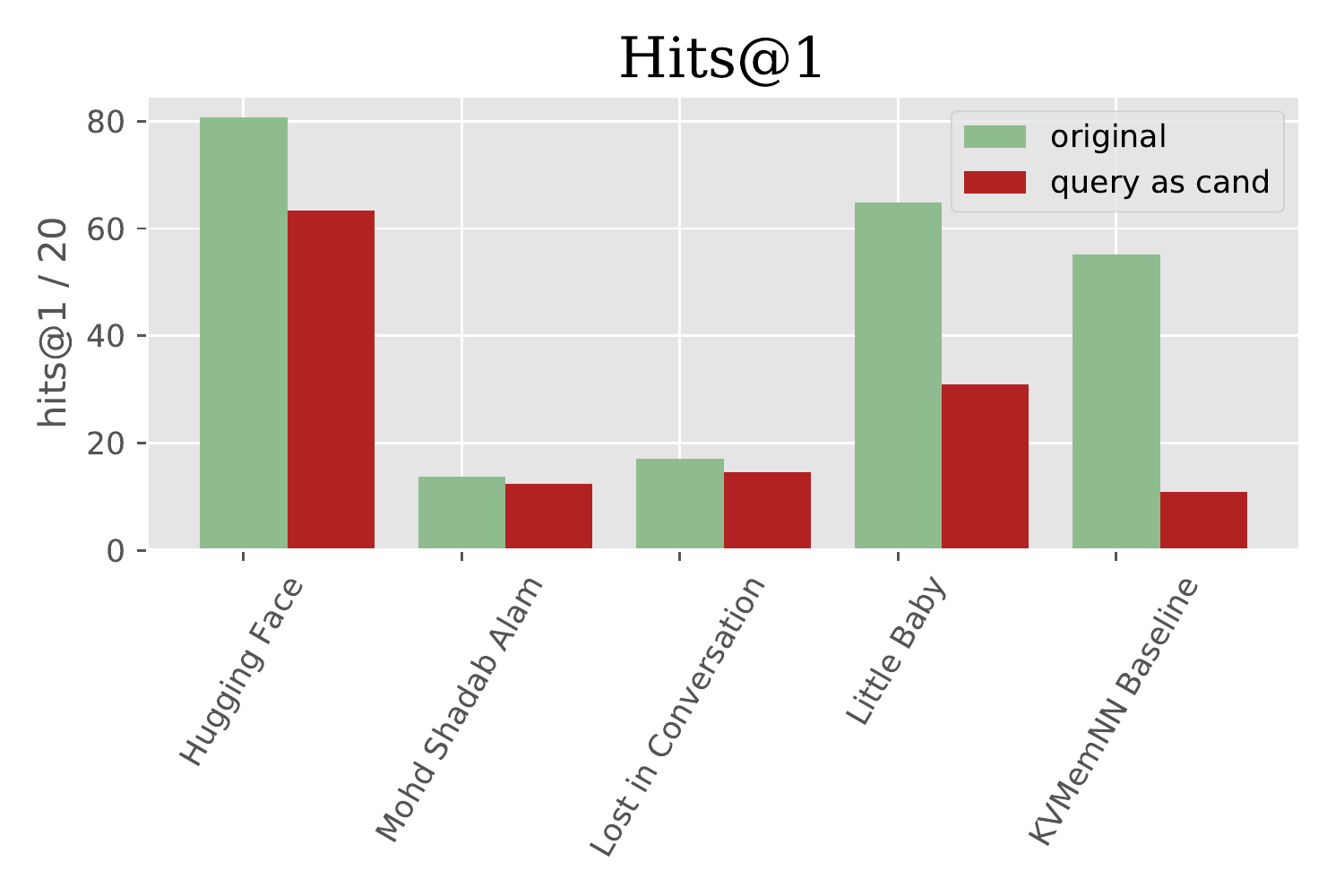}
    \caption{{\bf Distractor Candidate.} Hits@1 on the test set when we add the query (last partner message) as a candidate.}
    \label{fig:distract}
\end{figure}

\paragraph{Last Utterance (Parrot) Distractor}
We also evaluated how adding a distractor candidate affected the performance of these ranking models. Namely, we added the last partner message to the list of candidates to rank. A model should only in very rare circumstances parrot the speaking partner, so the Hits@1 metric should remain at a similar score with and without this distractor.
See Figure \ref{fig:distract} for the results. Most models suffered with this metric, showing they probably rely too much on word overlap with the last utterance when performing ranking (generally a response does have word overlap with the last utterance, but still it should not be a copy -- this makes this a somewhat difficult function for models to learn). The Hugging Face model was the most resistant to this type of attack, but still suffered to some degree.

\subsubsection{F1 Metric Toy Baseline}\label{sec:f1}
During the automatic evaluation stage of the competition, we discovered that always replying with \emph{``i am you to do and your is like"} would outperform the F1 score of all the models in the competition. This toy baseline was constructed simply by picking several frequent words from the training set. Specifically, always replying with this message gives an F1 score of 19.6 on the test set and 20.5 on the validation set (compare to Hugging Face's scores of 19.5 and 19.1 on the test and validation sets respectively). In \cite{liu_evaluate}, the authors showed that word overlap metrics do not correlate well with human judgment for dialogue response generation systems. This is another indication that we should reconsider using F1 to evaluate our dialogue systems.

\subsection{Human Evaluations}
\subsubsection{Wild Evaluation}
Recall that we solicited volunteers to chat with and evaluate the top-performing models through the Facebook Messenger and Telegram APIs. This setup is closer to real-world chatbot applications than the Mechanical Turk evaluation set-up. As volunteers -- unlike Turkers -- are not paid and therefore less likely to follow the instructions closely, we expected the distribution of these results to be quite different than those from the Mechanical Turk evaluation. Two notable (but expected) differences between the results of the Mechanical Turk and ``wild" evaluations are that the wild evaluation had relatively fewer good (i.e. long and consistent) dialogues, and Mechanical Turk workers tended to rate the models higher than the volunteers in the wild evaluation. 
A detailed description of our ``wild" dataset and a comparison with the data from the first ConvAI competition will be described in detail in separate publication\footnote{ConvAI2 dataset of non-goal-oriented human-to-bot dialogues (2019). V. Logacheva, V. Malykh, A. Litinsky, M. Burtsev}.

Given that many  of the dialogues collected during the wild evaluation were either senseless, offensive, or simply not in line with the instructions, and we did not have a systematic way of distinguishing these from the ``good” dialogues, we decided to discount the results from the wild evaluation altogether. See Table \ref{table:wild_examples} for example conversations.  The data collected from this evaluation is available in the competition repo\footnote{\url{http://github.com/DeepPavlov/convai/data}}. Detecting this type of undesirable behavior remains an open (but we believe solvable) problem for evaluation of models in a live system.


\begin{table}[t]
\begin{center}
\begin{footnotesize}
\begin{tabular}{rl}
\toprule
{[HUMAN:]} & Ok. Bye \\
{[BOT:]} & Bye. \\
{[HUMAN:]} & Bye\\
{[HUMAN:]} & Bye \\
{[HUMAN:]} & Yo \\
{[BOT:]} & What video games?.\\
{[HUMAN:]} & Bye \\
{[BOT:]} & OK. \\
{[HUMAN:]} & Bye \\
{[HUMAN:]} & Bye \\
\midrule
{[\textbf{\emph{Score}}:]} & 5/5\\
\end{tabular}
\quad
\begin{tabular}{rl}
\toprule
{[HUMAN:]} & Hey! \\
{[BOT:]} & hello there how are you? do you like dogs? \\
{[HUMAN:]} & Good ! I like dogs, I grew up in a farm \\
{[BOT:]} & that is cool. i've a german shepard. \\
{[HUMAN:]} & what is your dog's name ? \\
{[BOT:]} & his name is pepper. do you have any pets? \\
{[HUMAN:]} & I used to have a dog but he died..... \\
{[BOT:]} & that is so sad! do you watch nhl hockey? \\
{[HUMAN:]} & no... not a big fan of sport ? \\
{[BOT:]} & no, i love watching it. i love the nhl \\
\midrule
{[\textbf{\emph{Score}}:]} & 4/5\\
\end{tabular}
\end{footnotesize}
  \caption{Example of a bad conversation (left) and a good conversation (right)
  from the wild evaluation. The model received a score of 5 for the spam conversation, and 4 for the good conversation.
    \label{table:wild_examples}
    }
\end{center} 
\end{table}

\subsubsection{Mechanical Turk Evaluation}
Since the results of the wild evaluation were ultimately discounted, the winner of the human evaluation round -- and therefore the winner of the competition -- was determined by performance in the Mechanical Turk Evaluation. As announced at the NeurIPS Competition Track Workshop, the Lost in Conversation team won the competition. 

\begin{figure}
    \centering
    \includegraphics[width=1.0\textwidth]{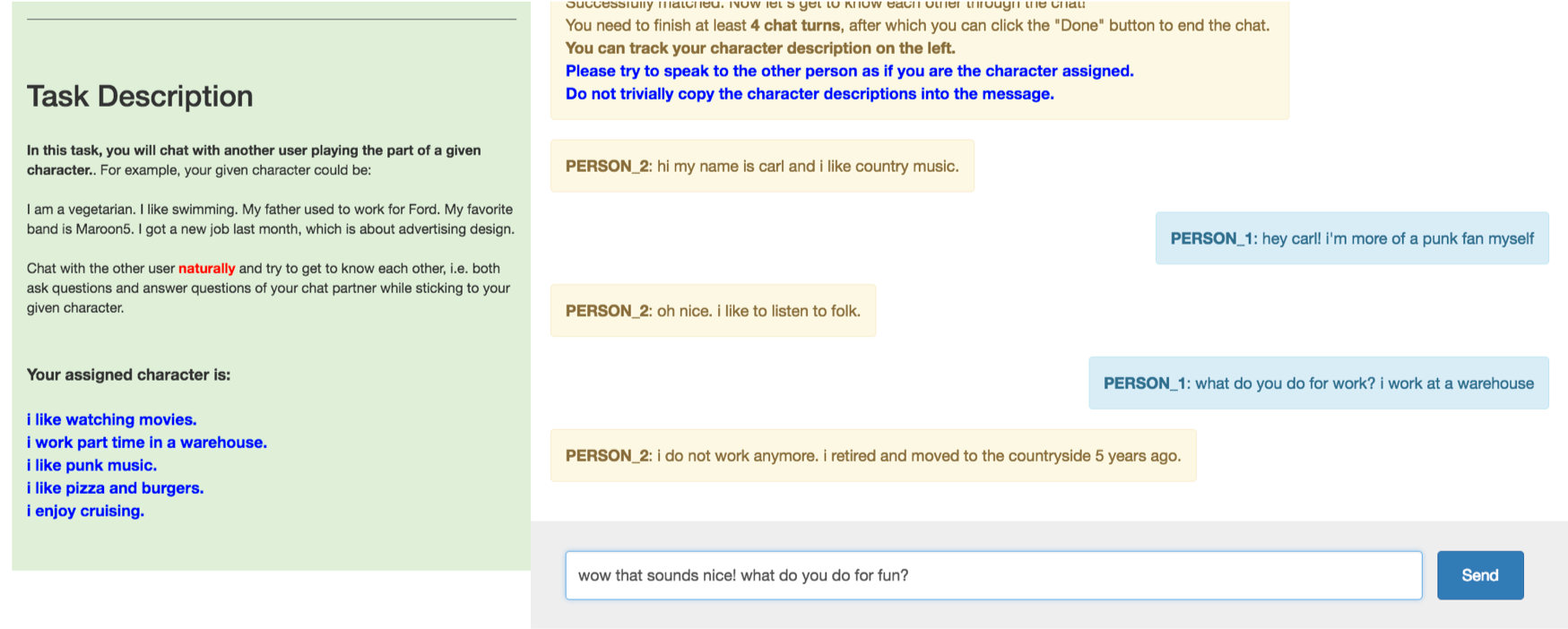}
    \caption{{\bf Mechanical Turk Evaluation Interface.} The chat interface used for the Mechanical Turk portion of the evaluation was intentionally similar to the interfae used to collect the original dataset.}
    \label{fig:mturksetup}
\end{figure}

The set-up of the Mechanical Turk evaluation was nearly identical to the set-up we used to collect the original {\sc Persona-Chat} dataset. The chat interface is shown in Figure \ref{fig:mturksetup}. For each evaluation, we paired a human worker with a model, assigned each of them personas, and instructed the humans to chat with and get to know their partner. Dialogues were of length 4-6 turns each. Following a short conversation, we asked workers ``How much did you enjoy talking to this user?" and had them answer on a scale of 1-4. Additionally, we tested whether the human could distinguish the persona the model was using from a random one. We crowdsourced 100 evaluations for each model.  Samples conversations from some of the models are given in Appendix \ref{app}.

\begin{table}[t]
\small
\begin{center}
\begin{tabular}{lcc}
\toprule
 {\bf Team Names}    & Engagingness (1-4) & Persona Detection (0-1)  \\
\midrule
1. Lost in Conversation & 3.11 & 0.9 \\
2. Hugging Face & 2.68 & 0.98 \\ 
3. Little Baby & 2.44 & 0.79  \\
4. Mohd Shadab Alam & 2.33 & 0.93 \\
5. Happy Minions & 1.92 & 0.46  \\
6. ADAPT Centre & 1.6 & 0.93  \\
\midrule 
Human & 3.48 & 0.96 \\
KV Profile Memory (Baseline) & 2.44 & 0.76 
\end{tabular}
\end{center}
\caption{Human Evaluation Results}
\label{humanleaderboard}
\end{table}

\begin{figure}
    \includegraphics[width=0.5\textwidth]{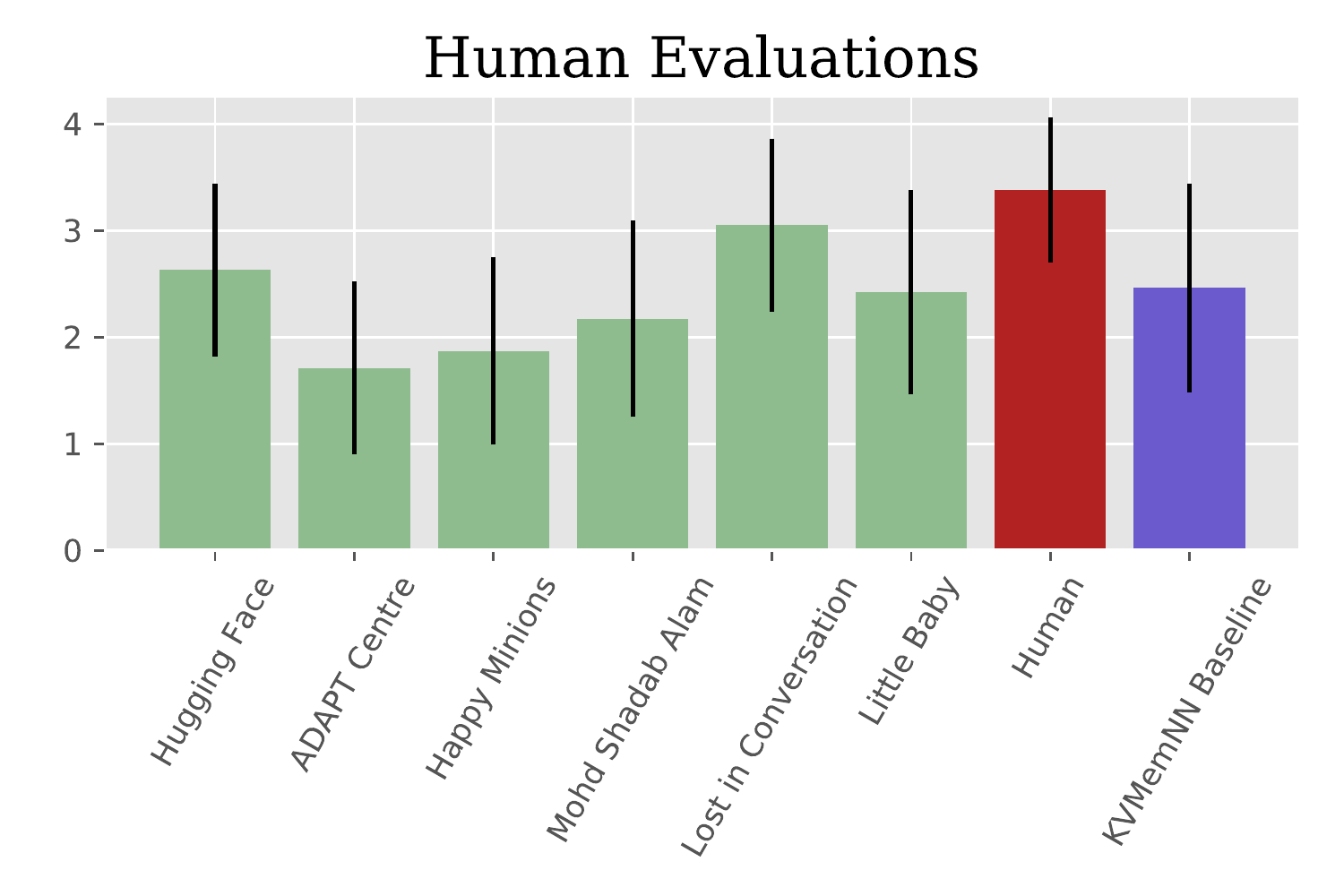}
    \includegraphics[width=0.5\textwidth]{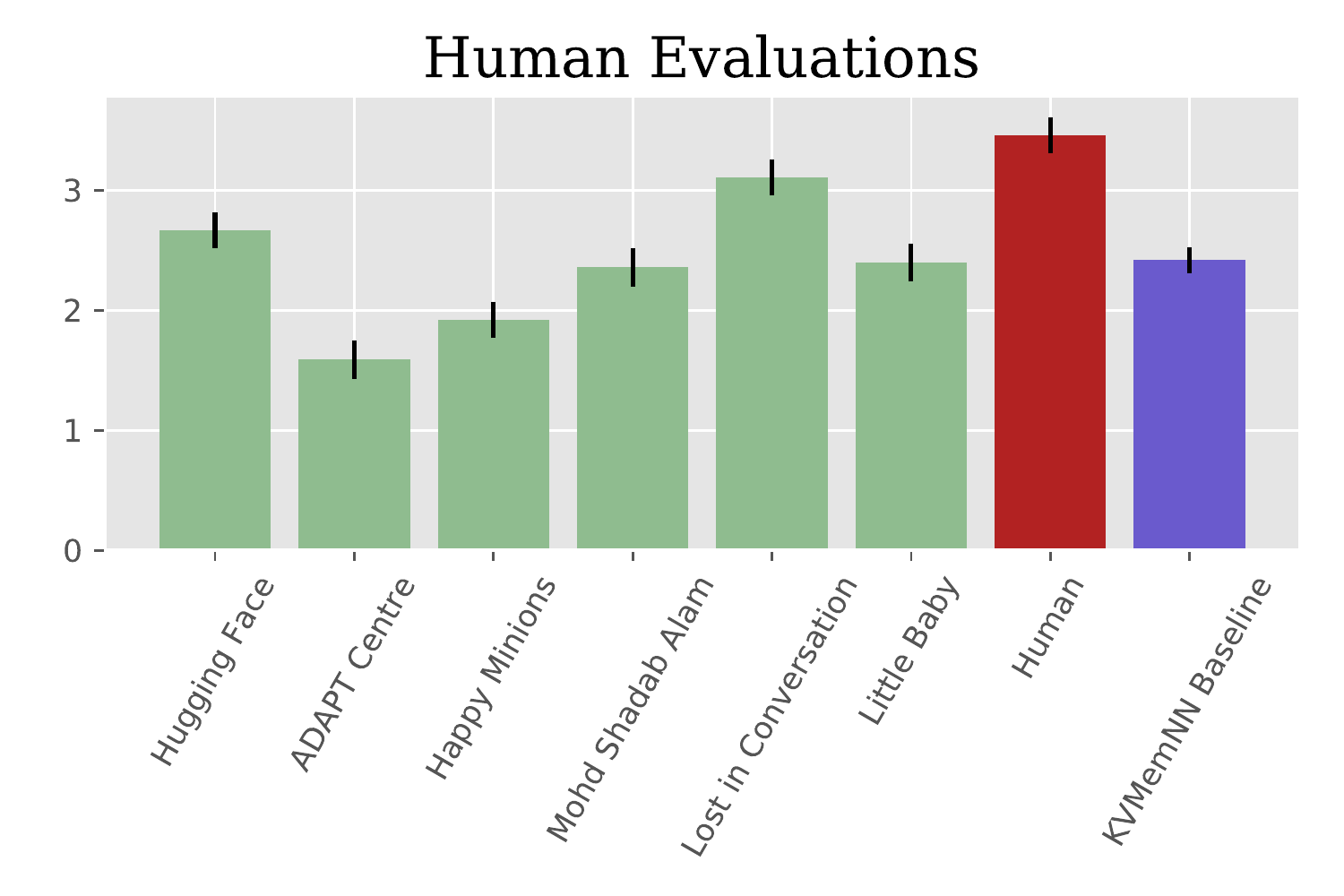}
    \caption{{\bf Mechanical Turk Evaluation: Engagingness.} Results before (left) and after (right) Bayesian calibration. The calibration did not alter the ordering of the scores.}
    \label{fig:engagingness}
\end{figure}

The results are shown in Table \ref{humanleaderboard}. Lost in Conversation won the competition with an engagingness score of 3.11 out of 4. We attempted to reduce annotator bias in the engagingness scores by using a Bayesian calibration method recently proposed in \cite{ilya}. The results from before and after calibration are given in Figure \ref{fig:engagingness}. The calibration did not affect the ordering of the scores, and the scores reported in the final leaderboard are post-calibration.

\subsubsection{Further Analysis of Results}\label{sec:furthur}
\paragraph{Length Statistics} In an attempt to understand the results from the Mechanical Turk evaluations, we analyzed various word statistics on the conversation logs. We measured the average length of both the bot and human responses for each team's evaluation, as shown in Table \ref{responselength}. Models with higher evaluation scores tended to get longer responses from humans, which can be considered as an implicit engagement score. 
However, this is possibly skewed by humans mimicking the length of the bot's utterances, e.g. consider ADAPT Centre's results. We note that when humans are speaking with other humans, they have much longer utterances on average than the models do. We believe this is related to their production of more generic, less engaging utterances.

\begin{table}[t]
\small
\begin{center}
\begin{tabular}{lccccccc}
\toprule
 {\bf Team Names}    & Engagingness & \# words & \# words & \# chars & \# chars  \\
  & (1-4) & (model) & (human) & (model) & (human) \\ 
\midrule
1. Lost in Conversation & 3.11 & 10.18 & 11.9 & 39.2 & 48.2 \\
2. Hugging Face & 2.67 & 11.5 & 11.9 & 44.4 & 49.2 \\
3. Little Baby & 2.4 & 11.5 & 11.3 & 51.5 & 47.3 \\
4. Mohd Shadab Alam & 2.36 & 9.5 & 10.2 & 33.8 & 42.5 \\
5. Happy Minions & 1.92 & 8.0 & 10.2 & 27.9 & 42.5 \\
6. ADAPT Centre & 1.59 & 15.1 & 11.8 & 60.0 & 48.0 \\
\midrule 
Human & 3.46  & -  & 13.7 & - & 57.7
\end{tabular}
\end{center}
\caption{{\bf Average response length in Mechanical Turk logs.} }
\label{responselength}
\end{table}

\paragraph{Rare Word Statistics}
We also looked to see how often rare words were used in the conversation logs. In Table \ref{rarewords}, Freq1h and Freq1k indicate the frequency with which the model used
words that appear fewer than 100 or 1000 times in the training corpus. 
The hypothesis here is that utterances with some rare words might be less generic and hence more interesting/engaging, rendering higher human evaluation scores.
The results show that humans use significantly more rare words than any of the models, and the bottom three models do have lower Freq1h scores than the top three; otherwise, however, the relationship between evaluation score of the models 
and their use of rare words is not completely clear. We suspect that is because this is just one factor among many that would need to be disentangled.

\begin{table}[t]
\small
\begin{center}
\begin{tabular}{lccccccc}
\toprule
 {\bf Team Names}    & Engagingness & Freq1h & Freq1h & Freq1k & Freq1k \\
  & (1-4) & (model) & (human) & (model) & (human) \\ 
\midrule
1. Lost in Conversation & 3.11 & 2.2 & 3.4 & 9.9 & 13.2 \\
2. Hugging Face & 2.67 & 2.5 & 4.2 & 9.0 & 15.6 \\
3. Little Baby & 2.4 & 4.9 & 3.7 & 18.3 & 15.6 \\
4. Mohd Shadab Alam & 2.36 & 1.3 & 3.2 & 9.5 & 14.1 \\
5. Happy Minions & 1.92 & 0.3 & 4.1 & 4.3 & 14.3 \\
6. ADAPT Centre & 1.59 & 1.7 & 3.5 & 8.8 & 15.1 \\
\midrule 
Human & 3.46  & 4.8 & 4.3 & 17.2 & 16.3
\end{tabular}
\end{center}
\caption{{\bf Rare word frequencies in Mechanical Turk logs.} }
\label{rarewords}
\end{table}

\paragraph{Word and Utterance Repetition Statistics}
We then looked at how often the models repeated themselves in conversations with humans. Table \ref{repeats} shows the frequency of unigram, bigram, and trigram repeats in the model responses, as well as how often the model's responses were unique in the logs. Again, it is clear the humans repeat themselves very infrequently, but there is not a clear relationship between our proxy measures of repetition with the human evaluation scores.
We suspect this is because there are more subtle instances of repeating that our proxies do not measure, and the proxies have already been optimized by many models (e.g. by doing $n$-gram or full utterance blocking).
For example we observed instances like ``i like watching horror'' followed by ``i love watching scary movies'' occurring, but these are not captured well by our metrics. Finally, overall utterance uniqueness should ideally be close to 100\% with the same utterance rarely being repeated across conversations, with humans at 99\%. While Hugging Face's model was at 97\%, many other models were lower, with the winner Lost in Conversation at 86\%.
A low uniqueness score could be problematic for a deployed system, as it might make users tire of it repeating itself. However, as our competition evaluations involve very short dialogues, this likely did not impact human evaluations.

\begin{table}[t]
\small
\begin{center}
\begin{tabular}{lccccccc}
\toprule
 {\bf Team Names}    & Engagingness & Unigram & Bigram & Trigram & Unique \\
  & (1-4) & Repeats & Repeats & Repeats & Responses \\ 
\midrule
1. Lost in Conversation & 3.11 & 2.11 & 5.6 & 2.67 & 86\%\\
2. Hugging Face & 2.67 & 1.49 & 5.04 & 0.6 & 97\%\\
3. Little Baby & 2.4 & 2.53 & 2.69 & 1.43 & 91\% \\
4. Mohd Shadab Alam & 2.36 & 3.48 & 11.34 & 7.06 & 83\% \\
5. Happy Minions & 1.92 & 1.62 & 6.56 & 3.81 & 53\% \\
6. ADAPT Centre & 1.59 & 6.74 & 11.53 & 1.44 & 98\% \\
\midrule 
Human & 3.46  & 1.83 & 2.47 & 0.51 & 99\%
\end{tabular}
\end{center}
\caption{{\bf Repeats in Mechanical Turk logs.} }
\label{repeats}
\end{table}

\paragraph{Blind Evaluation} Following the above analyses, it was still unclear why the Lost in Conversation model had a statistically significant human evaluation win over the Hugging Face model, even though the Hugging Face model performed much better in the automatic evaluations. To better understand this, we performed a blind evaluation ourselves of a random sample of the Mechanical Turk evaluation logs from these two teams, giving each conversation a score between 1 and 4 and making comments about the model's performance. The average score given to this subset of conversations is shown in Table \ref{blind}. As you can see, despite the apparent annotator bias, each annotator agreed with the Turkers regarding which model was better.

\begin{table}[t]
\small
\begin{center}
\begin{tabular}{lcc}
\toprule
    & \textbf{Hugging Face} & \textbf{Lost in Conversation}  \\
    \midrule
 \textbf{Turker} & 2.8 & 3.29 \\
 \textbf{Blind Annotator 1} & 2.47 & 2.78 \\
 \textbf{Blind Annotator 2} & 2 & 2.71 \\
\midrule
\end{tabular}
\end{center}
\caption{\textbf{Blind Evaluation Results.} Average engagingness score (1-4) for the randomly sampled subset of conversations.}
\label{blind}
\end{table}

\paragraph{Asking questions} Reading through the comments made by the blind annotators afterwards,  we noticed that while both models suffered from errors involving repetition, consistency or being ``boring"' at times, a common complaint about the Hugging Face model was that it ``asked too many questions." In order to determine to what extent this was true, we analyzed the Mechanical Turk logs and measured how often each model response began with a question word (like ``who," ``what," ``when," ``where," ``why," or ``how") and how often the response contained a question mark. 

\begin{figure}
    \includegraphics[width=0.5\textwidth]{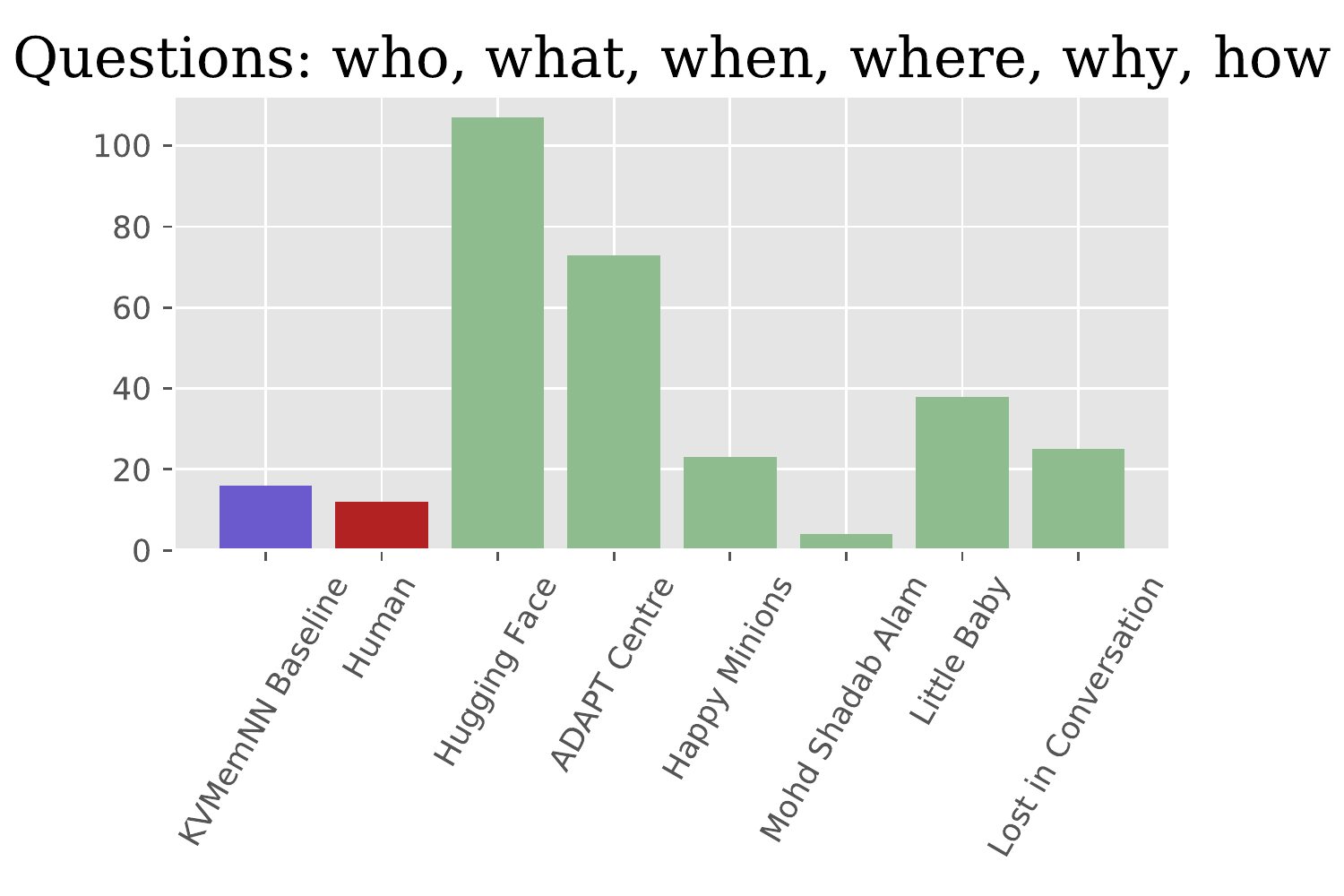}
    \includegraphics[width=0.5\textwidth]{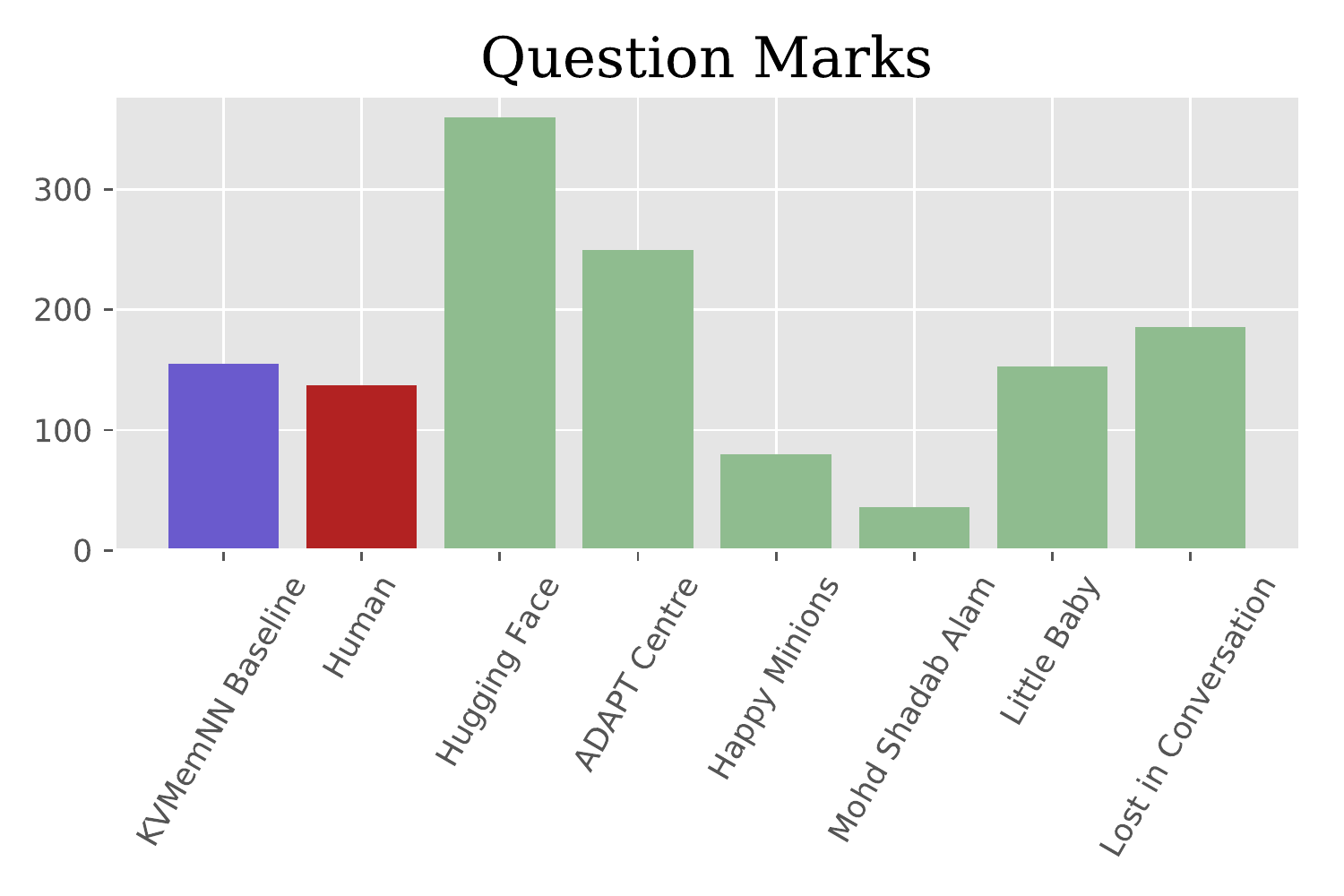}
    \caption{{\bf How often did the models ask questions?} We measured (on the left) how often the models began their response with ``who," ``what," ``when," ``where," ``why," or ``how," as well as (on the right) how often the models' responses contained at least one question mark as an estimate for how often the models asked questions when conversing with humans.}
    \label{fig:questions}
\end{figure}

The results are given in Figure \ref{fig:questions}. It is clear that the Hugging Face model is indeed a large outlier. Notably, you can see that in the 100 conversations it had, it began a response with a question word 107 times whereas humans only did this 12 times. When the model asks too many questions it can make the conversation feel disjointed, especially if the questions do not relate to the previous conversation. Friendly chit-chat requires a delicate balance of question-asking and question-answering. The tentative conclusion that we draw here is that the tendency to ask too many questions negatively affected the human evaluation results for the Hugging Face model. Future work should consider how we can automatically evaluate this type of conversation-level performance rather than just utterance-level performance.

\paragraph{Persona Detection} Lastly, looking at the persona detection scores from the Mechanical Turk evaluation in Table \ref{humanleaderboard}, we note that most models did relatively well in this metric (with the exception of the Happy Minions model). Recall that this score is the percentage of the time that the annotaters were able to to distinguish the model's persona from a random one.  We often observed models repeating the persona sentences almost verbatim, which might lead to a high persona detection score but a low engagingness score. 
Training models to use the persona to create engaging responses rather than simply copying it remains an open problem.

\section{Participating Models} \label{participants}

We include a short summary of the model types used for some of the top
competitors in Table \ref{modeltypes}.
Some of the authors of these models plan to write detailed papers describing their models. Please also refer to the slides at the website written by the model's authors\footnote{\url{http://convai.io/NeurIPSParticipantSlides.pptx}}.
The winner's (Lost in Conversation's) code is also publicly available\footnote{\url{https://github.com/atselousov/transformer_chatbot}}.

\begin{table}[h!]
\small
\begin{center}
\begin{tabular}{ll}
\toprule
 {\bf Team Names}    &  Model Summary \\ 
\midrule
Lost in Conversation &  Generative Transformer based on OpenAI GPT. Trained on\\ 
&  {\sc Persona-Chat} (original+revised), DailyDialog and Reddit comments. \\
\hline
Hugging Face &  Pretrained generative Transformer (Billion Words  + CoNLL 2012)\\ & with transfer to {\sc Persona-Chat}. \\
\hline
Little Baby & Profile-Encoded Multi-Turn Response Selection  \\
& via Multi-Grained Deep Match Network. \\
&Modification of \cite{wu2016sequential}: better model + data augmentation via translation.\\
\hline
Mohd Shadab Alam & Seq2Seq + Highway model.\\ 
& Glove + language model vector.\\
& Transfer learning strategy for Seq2Seq tasks.\\
\hline
ADAPT Centre&  Bi-directional Attentive LSTM. \\
            & Pretrained via GloVe embeddings + Switchboard, Open Subtitles.
\end{tabular}
\end{center}
\caption{{\bf Brief model descriptions of some of the top competitors.} }
\label{modeltypes}
\end{table}

\section{Conclusions and Future Work}

\paragraph{Models}
The best models in the competition were variants of the generative Transformer architecture.  Those models have rather high capacity and thus cannot be trained on ConvAI2 ({\sc Persona-Chat})  data alone, but must be either pretrained or multitasked with additional large datasets. One can use dialogue datasets to pretrain, but it seems as though the system still works well with language modeling datasets that are not explicitly dialogue (e.g. the Billion Words corpus). Many other tweaks to the base models were tried, such as trying to optimize the automatic metrics directly, but without direct ablations with human evaluation it is difficult to state here the effects of all these components.

Retrieval models fared a little worse than generative models in the human evaluations, although we are unsure if this is true in general, or because no very strong retrieval model was proposed. With a Transformer-based retrieval model it is possible to get Hits@1 in excess of 80\% but no such method was tried by a competitor (see Table \ref{automaticleaderboard}, Hugging Face used a two-head Transformer model, but opted to generate rather than retrieve). 
In our opinion, looking at the outputs from the generative systems in the competition, they still fall short of the most interesting and engaging comments of humans (which sometimes retrieval models choose); however, the generic responses from generative models are often low-risk or ``safe" responses, which may give them higher scores. A retrieve and refine approach (combining generative and retrieval methods) is another possibility that was not explored in the competition \cite{weston2018retrieve}.

Finally, better sentence representations are being developed all the time. This competition was run before the BERT model \cite{devlin2018bert} was released which has been shown to improve many NLP tasks. Hence, we expect these models to improve on ConvAI2 as well.

\paragraph{Automatic vs. Human Evaluation}
It remains an open problem to find the best automatic evaluation metrics for dialogue.
There is not enough data from the competition to measure correlation between the automatic metrics we tried and human evaluations in depth. 
%
Clearly a randomly initialized model has poor values for all of these metrics, whereas training to optimize any of them will improve human evaluations. The problem is more whether the finer-grained differentiation of relatively similar models can be automatically measured.
We believe each automatic metric evaluates at least some aspects of what humans consider a ``good" model but misses other aspects. As such, optimizing only one of these metrics can fail to address important 
issues.  For example, optimizing per-word 
perplexity fails to address the search strategy of a model when generating a full utterance, e.g. it is not affected by beam search choices. Optimizing Hits@1 is a per-utterance metric that fails to address the full conversational flow (as the gold dialogue history between two humans is used for that metric, not what the model previously said). Some models optimize F1 and do well, however it  also has major issues
(see Section \ref{sec:f1}). 
Further, it is very  hard to compare retrieval and generative models other than by human evaluation.

Nevertheless, we find the use of automatic metrics important for several reasons.
If we desire to be able to train our models offline at least initially (which we believe we do) then we need an offline training objective, which typically relates to automatic metrics. Hence, if we understand how human evaluations relate to automatic metrics, not only will we understand the dialogue task better, but we will know how to perform such offline training. Additionally, for our competition it would have been very difficult to filter models for the human evaluation stage without the use of automatic metrics.

\paragraph{Towards Multi-turn Evaluation} We thus believe we are still 
missing some key offline (automatic) metrics, but have hope that they are possible to find. We identified that the current metrics fail to measure the multi-turn aspects of human evaluation, in particular in terms of repetition, consistency and balance of dialogue acts. 
Even the best competitors' models often failed to be self-consistent across a few dialogue turns, 
which we believe was at least partly responsible for  lowering their evaluation score. For example, ``i am a professional runner. you? i love running'' followed by ``i'm not very athletic''  or ``i work as a snowboard instructor'' followed by ``i work for a food company'' are both unlikely continuations of a conversation. Even if they happen infrequently, these problems are particularly jarring for a human speaking partner when they do happen.

In a related problem, we observed the models asking questions that are already answered, e.g. one model asks ``what do you do for a living?'' even though the human earlier stated ``i work on computers'' resulting in the human replying ``I just told you silly''.

One possible solution to these problems is the use of dialogue natural language inference (NLI) \cite{welleck2018dialogue}, a new task that has been proposed that evaluates exactly these problems. It works by providing pairs of utterances as input, and the task is then to predict if the pair entail, are neutral or contradict.
This is exciting because it can allow us to both (i) fix our model's consistency problems by training on this new task and (ii) evaluate to what extent our model's consistency problems are fixed using the evaluation set.

Finally,  in Section \ref{sec:furthur} we identified that models that do not balance question asking with answering over multiple turns might can cause human evaluations to suffer.
Given this information, it may be possible to construct new metrics that 
measure these kind of balances so that we can optimize them (to look more similar to human data, for instance).

\paragraph{Towards more complex tasks}
Going forward, even if we can completely solve the ConvAI2 {\sc Persona-Chat} task (i.e. reach human performance), it is still only a meet-and-greet task involving getting to know someone for a few dialogue turns, with shallow topics and quick context switches. Clearly many aspects of an intelligent agent are not evaluated by this task, such as the use of long-term memory or in-depth knowledge and deeper reasoning.
For example, in Table \ref{table:persona-chat-example} ``Game of Thrones'' is mentioned, but a model imitating this conversation would
not really be required to know anything more about the show, as in ConvAI2 speakers tend to shallowly discuss each other's interest without lingering on a topic for too long.
Subsequent competitions could explore this issue further. Such a study is feasible as several new 
datasets are being released to explore such a setting, in particular
the Wizard of Wikipedia task involves using knowledge from Wikipedia to discuss open-domain topics \cite{dinan2018wizard}.
The DSTC7 competition\footnote{\url{http://workshop.colips.org/dstc7/}} also recently addressed this topic, 
however the evaluation was not multi-turn.

\section{Acknowledgements} 

We thank all the competitors for taking part and making this a successful competition.
We especially thank the competition's sponsors, Facebook Academics and 
Amazon Web Services. Participation of Mikhail Burtsev, Varvara Logacheva, and Valentin Malykh was supported by National Technology Initiative and PAO Sberbank project ID 0000000007417F630002.

\bibliographystyle{unsrt}
\bibliography{iclr2018_conference}

\appendix

\section{Example Dialogues}\label{app}

Example dialogues for some of the top models are given
in Figures \ref{figure:lostinconvo}, \ref{figure:huggingfaceconvo}, \ref{figure:littlebabyconvo},  \ref{figure:mohdshadabalamconvo}, \ref{figure:happyminionsconvo}, and \ref{figure:adaptcentreconvo}.

\begin{figure}[h]
\small
\begin{center}
\begin{tabular}{rl}
\toprule
{[BOT:]} & hello, how are you today? \\
{[HUMAN:]} & I'm great!  I just got back from skydiving! I love doing that! \\
{[BOT:]} & that sounds like fun! i love to paint.  \\
{[HUMAN:]} & Awesome!  What do you like to paint? \\
{[BOT:]} & i love to draw portraits of my favorite cars. \\
{[HUMAN:]} & I'd love to see them!  I live in Arizona, where do you live? \\
{[BOT:]} & i live in california. i love the beach! \\
{[HUMAN:]} & I've been to California!  I love to shop at their Barnes and Noble store! \\
{[BOT:]} & i love shopping at kroger! \\
{[HUMAN:]} & Kroger is great!  I wish they sold pet food for my pet lizard though! \\
\midrule
{[\textbf{\emph{Score}}:]} & 4/4\\
\end{tabular}
  \caption{Lost in Conversation: example Mechanical Turk conversation.
    \label{figure:lostinconvo}
    }
\end{center} 
\end{figure}

\begin{figure}[h]
\small
\begin{center}
\begin{tabular}{rl}
\toprule
{[HUMAN:]} & Hello there, I like rock climbing do you have any hobbies? \\
{[BOT:]} & i like nature. do you? i like nature. \\
{[HUMAN:]} & I do love nature, especially when going on walks \\
{[BOT:]} & what is your favorite rock? mine is rock.  \\
{[HUMAN:]} & Well I like rock music, does that count? \\
{[BOT:]} & i think it does. what do you do for fun? \\
{[HUMAN:]} & Well I work at a restaurant that I love, and rock climbing \\
{[BOT:]} & that sounds fun! i'm into nature. \\
{[HUMAN:]} & Cool, so am I \\
{[BOT:]} & do you have any pets? do you have pets? \\
\midrule
{[\textbf{\emph{Score}}:]} & 2/4\\
\end{tabular}
  \caption{Hugging Face: example Mechanical Turk conversation.
    \label{figure:huggingfaceconvo}
    }
\end{center} 
\end{figure}

\begin{figure}[h]
\small
\begin{center}
\begin{tabular}{rl}
\toprule
{[BOT:]} & whats its name, she good off leash? \\
{[HUMAN:]} & I don't actually have any dogs. \\
{[BOT:]} & its name is noodle   she does not need a leash   i talk to her in my sleep  \\
{[HUMAN:]} & Wow! Does she talk back \\
{[BOT:]} & why do you not believe her \\
{[HUMAN:]} & I don't even know what she said. \\
{[BOT:]} & after my workouts i drink a protein shake instead coffee, do not like coffee \\
{[HUMAN:]} & Alright, that's cool. \\
\midrule
{[\textbf{\emph{Score}}:]} & 1/4\\
\end{tabular}
  \caption{Little Baby: example Mechanical Turk conversation.
    \label{figure:littlebabyconvo}
    }
\end{center} 
\end{figure}

\begin{figure}[h]
\small
\begin{center}
\begin{tabular}{rl}
\toprule
{[BOT:]} & hi how are you today? \\
{[HUMAN:]} & Hello how are you? \\
{[BOT:]} & i'm good. just got done reading a book. \\
{[HUMAN:]} & Which book did you read? \\
{[BOT:]} & i don't read much, i am more into reading \\
{[HUMAN:]} & Where do you live? \\
{[BOT:]} & i live in the us. i love to read \\
{[HUMAN:]} & I live in Alaska \\
{[BOT:]} & i live in the us, i love to read \\
{[HUMAN:]} & Do you like to read? \\
\midrule
{[\textbf{\emph{Score}}:]} & 1/4\\
\end{tabular}
  \caption{Mohd Shadab Alam: example Mechanical Turk conversation.
    \label{figure:mohdshadabalamconvo}
    }
\end{center} 
\end{figure}

\begin{figure}[h]
\small
\begin{center}
\begin{tabular}{rl}
\toprule
{[BOT:]} & i love to read \\
{[HUMAN:]} & Nice, I'm more of a numbers guy myself. I use to be an accountant. \\
{[BOT:]} & well, i am not sure what you mean. \\
{[HUMAN:]} & I worked with numbers. I recently moved away to a beach house because I got tired of it \\
{[BOT:]} & i am sorry to hear that. do you have any hobbies? \\
{[HUMAN:]} & Just hanging out with my dog, I guess. Trying to start a new business. \\
{[BOT:]} & i am not sure what you mean. \\
{[HUMAN:]} & Exactly, it's hard to find meaning \\
{[BOT:]} & yeah, i guess. i am not sure what that means. \\
{[HUMAN:]} & I do work, for pay, occasionally, as a business person. \\
\midrule
{[\textbf{\emph{Score}}:]} & 1/4\\
\end{tabular}
  \caption{Happy Minions: example Mechanical Turk conversation.
    \label{figure:happyminionsconvo}
    }
\end{center} 
\end{figure}

\begin{figure}[h]
\small
\begin{center}
\begin{tabular}{rl}
\toprule
{[HUMAN:]} & Hello, have you ever gone scuba diving? \\
{[BOT:]} & no i do not have a job yesterday. do you have any hobbies \\
{[HUMAN:]} & I love going scuba diving in hawaii. \\
{[BOT:]} & i love scuba diving yesterday. do you like to swim? i do \\
{[HUMAN:]} & It's ok, I prefer to play with my pet snakes. \\
{[BOT:]} & i love to swim yesterday. i like to go to the beach with my job \\
{[HUMAN:]} & I just got divorced so maybe I should go to the beach. \\
{[BOT:]} & you should go with your job yesterday. do you go to a lot of the beach \\
{[HUMAN:]} & Only when I visit hawaii. \\
{[BOT:]} & i love the beach so i have to go with my new job \\
{[HUMAN:]} & I traveled the Americas playing guitar \\
\midrule
{[\textbf{\emph{Score}}:]} & 2/4\\
\end{tabular}
  \caption{ADAPT Centre: example Mechanical Turk conversation.
    \label{figure:adaptcentreconvo}
    }
\end{center} 
\end{figure}

\end{document}